\documentclass[11pt,a4paper]{article}
\usepackage[hyperref]{acl2019}
\usepackage{times}
\usepackage{latexsym}
\usepackage{CJKutf8}
\usepackage{multirow}
\usepackage{multicol}

\usepackage{algorithm}
\usepackage{algorithmic}

\usepackage{url}

\usepackage{bbm}
\usepackage{graphicx}
\usepackage{booktabs} 
\usepackage{makecell}

\graphicspath{ {.} }

\aclfinalcopy 
\def\aclpaperid{295} 

\title{Machine Translation in Pronunciation Space}
\author{Hairong Liu$^1$ \,
Mingbo Ma$^1$ \,
Liang Huang$^{1,2}$ \,
\\[0.1cm]
  $^{1}$Baidu Research, Sunnyvale, CA, USA
\\
  $^2$Oregon State University, Corvallis, OR, USA
\\
  {\tt \small \{liuhairong, mingboma,lianghuang\}@baidu.com }
}

\date{}

\begin{document}
\maketitle
\begin{CJK}{UTF8}{gbsn}
\begin{abstract}
The research in machine translation community focus on translation in text space. However, humans are in fact also good at direct translation in pronunciation space. Some existing translation systems, such as simultaneous machine translation, are inherently more natural and thus potentially more robust by directly translating in pronunciation space. In this paper, we conduct large scale experiments on a self-built dataset with about $20$M En-Zh pairs of text sentences and corresponding pronunciation sentences. We proposed three new categories of translations: $1)$ translating a pronunciation sentence in source language into a pronunciation sentence in target language (P2P-Tran), $2)$ translating a text sentence in source language into a pronunciation sentence in target language (T2P-Tran), and $3)$ translating a pronunciation sentence in source language into a text sentence in target language (P2T-Tran), and compare them with traditional text translation (T2T-Tran). Our experiments clearly show that all $4$ categories of translations have comparable performances, with small and sometimes ignorable differences.
\end{abstract}

\section{Introduction}
Machine translation (MT) is a hot research topic in NLP community for about three decades \cite{brown1990statistical, cho2014learning, koehn2009statistical, bahdanau2014neural}, and its performance is steadily getting better and better. In recent years, due to the rapid progress of neural machine translation (NMT) \cite{bahdanau2014neural, luong2015effective, sutskever2014sequence}, the performance of many deployed MT systems, such as Google Translate \cite{wu2016google}, gradually approach human's performance. However, the research in machine translation community almost exclusively focus on text translation (T2T-Tran).

\begin{figure*}[ht]
\includegraphics[scale=0.36]{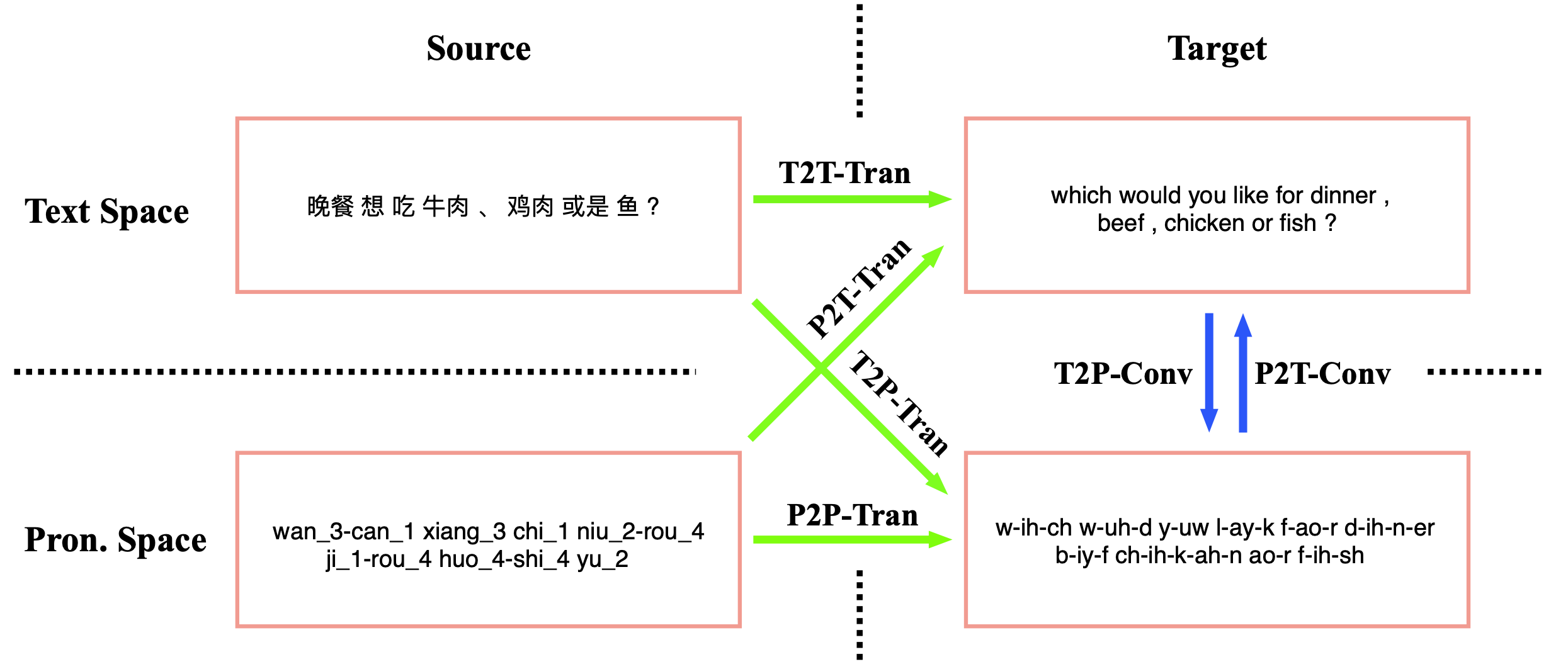}
\caption{The illustration of 4 categories of translations, namely, P2P-Tran, P2T-Tran, T2P-Tran and T2T-Tran, and 2 categories of conversion, namely, T2P-Conv and P2T-Conv. The pronunciation of the Chinese sentence is expressed as a sequence of Pinyins; while the pronunciation of the English sentence is expressed as a sequence of phonemes. The symbol `-' is used to separate Pinyins or phonemes in a word. Note that the punctuation information is lost in pronunciations. }
\label{fig:system}
\end{figure*}

For human communication, text is only a tool of recent era, and it only represents a small portion of human communication \cite{panini2015oral}. In a lot of scenarios of our daily life, communication occurs in pronunciation space. For example, an illiterate person can freely talk to others, and exchange information precisely. Translation also naturally occurs in pronunciation space. For example, a kid grew up in a multilingual family often involuntarily express the same thing in different languages in reaction to different targets. Thus, research on translation in pronunciation space is of great interest to us as a human being, as well as to NLP community.

Many existing translation systems may also benefit from translation in pronunciation space \cite{zheng2011chime, oda2014optimizing}. One such example is a simultaneous machine translation system (SMT) \cite{oda2014optimizing, grissom2014don, ma2019stacl}.
In SMT, the input of the translation module is the output of an automatic speech recognition module (ASR) \cite{povey2011kaldi}, and the direct output of an ASR is usually in pronunciation space \footnote{The output of ASR is usually phonemes or tri-phones, which is then converted to texts}. The translation module is connected to a Text-to-speech (TTS) module \cite{arik2017deep}, whose input is also in pronunciation space. Obviously, it is natural to directly translate in pronunciation space for an SMT system. By eliminating the process of converting a sentence from pronunciation space to text space, and converting the translation back from text space to pronunciation space, a lot of errors are avoided \cite{formiga2012dealing, ruiz2015phonetically}, such as the very common homophone noises \cite{liu2018robust}, and the effect of error propagation is also reduced \cite{belinkov2017synthetic}.

According to our best knowledge, there is no previous work on translation in pronunciation space, especially in the context of large scale datasets and neural machine translation. Many techniques developed for text translation should also be useful for translation in pronunciation space, such as seq2seq models \cite{sutskever2014sequence, vaswani2017attention}, subword units \cite{sennrich2016neural}, etc. However, pronunciation space also has some unique characteristics, such as limited pronunciation units and no punctuations. Since the topic of translation in pronunciation space is untouched, there is a lot of questions to be answered, such as $1)$ how good the seq2seq model is in pronunciation translation?, $2)$ what is the effect on translation quality due to the lack of punctuations?, and $3)$ is syllable better than phoneme for pronunciation translation?, to name just a few.

To answer these questions and to verify the performance of translations in pronunciation space, in this paper, we conduct large scale experiments for the Zh-En language pair. Translation in pronunciation space has $3$ major categories, namely, P2P-Tran, P2T-Tran and T2P-Tran, all of them are complementary to T2T-Tran, the traditional text translation. For the purpose of comparison, we also need to convert a text sentence into its corresponding pronunciation sentence in the same language (T2P-Conv), as well as converting a pronunciation sentence into its corresponding text sentence in the same language (P2T-Conv). Figure \ref{fig:system} illustrates these 4 categories of translations and 2 categories of conversion. 

According to our experiments, T2P-Conv has almost perfect performance since this task is trivial. P2T-Conv is a little harder since a pronunciation sentence contains less information than its corresponding text sentence, but the performance is still very good (with BLEU scores are larger than $90$). All $4$ categories of translation have similar performances, which means that we are restricted to T2T-Tran, and can freely choose most suitable categories of translation for our systems. 

\section{Dataset}
Since we cannot find parallel pronunciation datasets online, we build a large scale dataset based on WMT18 Zh-En dataset \footnote{\url{http://data.statmt.org/wmt18/translation-task/preprocessed/zh-en/}}.  

There are about $24.75$M parallel sentences in WMT18 Zh-En dataset. For each parallel sentence pair $(s, t)$ where $s$ is a Chinese text sentence and $t$ is an English text sentence, we try our best to generate corresponding pronunciation sentences for both $s$ and $t$, denoted by $s_p$ and $t_p$, respectively. In this conversion, each text word will be converted into its corresponding pronunciation word, expressed as a sequence of Pinyins (with tones) for Chinese and a sequence of phonemes for English, and punctuations are in general discarded.  If we can get both $s_p$ and $t_p$, we get a dataset entry, which is a quadruple, $(s, t, s_p, t_p)$. Otherwise, we simply ignore this pair of sentences since we do not know the pronunciations of some words in them. In this way, we get a dataset with $20628950$ entries. Table \ref{tab:entry_examples} illustrates two exemplar entries in this dataset. We randomly split this dataset into \emph{train}, \emph{dev} and \emph{test} sets, with $20620758$, $4096$, $4096$ entries, respectively \footnote{We will release this dataset in the future.}. 

In this dataset, the number of words is $742463$ for Chinese and $752312$ for English, and the number of total tokens is $401054208$ and $456209710$, respectively, for Chinese and English. There are $1485$ Pinyins and $39$ phonemes, and the number of Chinese characters is $9536$.

The following subsection describes in detail how we generate the corresponding pronunciation sentence for a text sentence, for both Chinese and English.

\begin{table*}[ht]
\begin{center}
 \begin{tabular}{ll} \toprule
   $s$ & 2005 年 1 月 31 日   \\
   $t$ & 31 January 2005 \\
   $s_p$ & er\_4-ling\_2-ling\_2-wu\_3 nian\_2 yi\_1 yue\_4 san\_1-shi\_2-yi\_1 ri\_4 \\
   $t_p$ & th-er-d-iy-w-ah-n jh-ae-n-y-uw-eh-r-iy t-uw-th-aw-z-ah-n-d-ah-n-d-f-ay-v \\
   \midrule  
   $s$ & 晚餐 想 吃 牛肉 、 鸡肉 或是 鱼 ?   \\
   $t$ & which would you like for dinner , beef , chicken or fish ? \\                                      
    $s_p$ & wan\_3-can\_1 xiang\_3 chi\_1 niu\_2-rou\_4 ji\_1-rou\_4 huo\_4-shi\_4 yu\_2  \\
   $t_p$ & w-ih-ch w-uh-d y-uw l-ay-k f-ao-r d-ih-n-er b-iy-f ch-ih-k-ah-n ao-r f-ih-sh \\
     \bottomrule
    \end{tabular}
    \caption{Two entries in our dataset. $s$: Chinese text sentence, $t$: English text sentence, $s_p$: Chinese pronunciation sentence, $t_p$: English pronunciation sentence.
    The symbol `-' is used to segment different pronunciation units (Pinyins or phonemes) in a word.}
    \label{tab:entry_examples}
  \end{center}
\end{table*}

\subsection{From a Text Sentence to a Pronunciation Sentence}
To generate the corresponding pronunciation sentence of a text sentence, we get the pronunciation of each word in the text sentence one by one, and concatenate them. However, there are many special cases to deal with, such as how to get pronunciations of digit numbers, how to process some special punctuations which affect pronunciation, and how to get pronunciations of words missing in the lexicon. 

\subsubsection{Lexicon}
For Chinese, we use a large lexicon from DaCiDian \footnote{\url{https://github.com/aishell-foundation/DaCiDian}} which maps $516596$ Chinese words to their pronunciations, expressed in Pinyins. This lexicon is large and a lot words in this lexicon do not appear in WMT18 Zh-En dataset. When looking up lexicon for pronunciations, the tricky part is how to deal with words with multiple pronunciations. For such a word, there are two situations. First, there is a unique pronunciation of this word according to the context. For example, the word `重' has two pronunciations, namely, `chong\_2' and `zhong\_4'; however, in the word `重庆', the pronunciation of `重' is `chong\_2', and in the word `重量', the pronunciation of `重' is `zhong\_4'. In this situation, there is no ambiguity. Second, there are still multiple pronunciations of this word when taking the context into account. In this case, we always use the first one   \footnote{We also tried another approach which randomly chooses one pronunciation if there are multiple pronunciations.  Both approaches achieve similar performance on P2T-Tran; however, the random approach has much worse performance in both P2P-Tran and T2P-Tran due to the randomness introduced in the side of target language.}. 

For English, we use the Voxforge English Lexicon \footnote{\url{http://www.repository.voxforge1.org/downloads/SpeechCorpus/Trunk/Lexicon/}} which maps $268996$ words into their pronunciations, expressed in phonemes.
If a word has multiple pronunciations, we also only use the first one.

\subsubsection{Number}
For digital numbers in Chinese sentences, we first convert them into corresponding Chinese text numbers,  such as `22' is converted into `二十二', using a tool called `pycnnum' \footnote{\url{https://pypi.org/project/pycnnum/}}. If there is a symbol  `-' before digit numbers, it is converted into the Chinese character `负'.  Text numbers will be converted into Pinyins by looking up Table \ref{tab:chinese_number}, character by character.

For digit numbers in English sentences, we first convert them into corresponding text representation in English, such as `22' is converted into `twenty two', using a package called \emph{num2words} \footnote{\url{https://pypi.org/project/number2words/}}, and then convert them into phonemes.

\begin{table}[h]
\begin{center}
\begin{tabular}{lc|cc|cc} \toprule
零 & ling\_2 &  一 & yi\_1 & 二 & er\_4\\
三 & san\_1 & 四 & si\_4 & 五 & wu\_2 \\
六 & liu\_4 &七 & qi\_1 &八 & ba\_1 \\
九 & jiu\_2 &十&shi\_2&百&bai\_3 \\
千&qian\_1 &万&wan\_4&亿&yi\_4 \\
兆&zhao\_4&京&jing\_1&点&dian\_3\\
\bottomrule
\end{tabular}
 \caption{Mapping from Chinese characters to Pinyins for numbers.}
   \label{tab:chinese_number}
 \end{center}
\end{table}

\subsubsection{G2P}
For words which is neither in the lexicon nor a number, we then try to get their pronunciations by Grapheme-to-Phoneme tools (G2P) \cite{rao2015grapheme}. G2P is very effective and it produces pronunciations of most word missing in the lexicon. 

For Chinese, the G2P tool we use is \emph{pinyin}\footnote{\url{https://pypi.org/project/pinyin/}}; while for English, we use a tool called \emph{g2p-seq2seq} \footnote{\url{https://github.com/cmusphinx/g2p-seq2seq}}.

\subsubsection{Punctuation}
In general, we discard the punctuation information when converting a text sentence into a corresponding pronunciation sentence. However, there are some punctuations which directly affect pronunciations, such as the comma symbol (`,') which separates numerals in numbers and the percentage symbol (`\%') after numbers. For these punctuations, we process these affected words to make sure the pronunciation can be correctly generated. For example, for the percentage symbol, in Chinese, we add the word `百分之' before numbers; while in English, we add the word `percentage' after numbers.

\section{Subword Units for Translation}
Using subword units for translation is now almost a standard procedure in machine translation \cite{sennrich2016neural}. It solves the problems of rare words and out-of-vocabluary (OOV) words and thus improves the performance. In our experiments, for text sentences, we use a tool named \emph{sentencepiece} \footnote{\url{https://github.com/google/sentencepiece}} to learn subword units through byte-pair-encoding algorithm (BPE) \cite{kunchukuttan2016learning}, as well as decompose a text sentence into a sequence of subwords. We cannot directly use \emph{sentencepiece}  for pronunciation sentences since both Pinyins and phonemes are not fixed-size bytes. To get subwords in pronunciation space, we use  \emph{sentencepiece} in the following way: $1)$ build a one-to-one mapping between pronunciation units (Pinyins for Chinese and phonemes for English) and Chinese characters, $2)$ convert pronunciation sentences into pseudo Chinese text sentences and do BPE using \emph{sentencepiece}, and $3)$ convert back into pronunciation sentences.

In English, vowels play an import role in segmenting pronunciations. The subwords learnt by  \emph{sentencepiece}  in English pronunciation space may contain zero, one or more vowels. Table \ref{tab:english-bpe} shows some examples. A well-known pronunciation unit in English is syllable, which contains only one vowel. Naturally, we want to know the performance of translation with syllables being basic units. Since there is no widely recognized approach to decompose the pronunciation of a word into syllables, we propose a modified BPE algorithm to achieve this goal.

\begin{table*}[h]
\begin{center}
\begin{tabular}{l|ccccc} \toprule
\#vowels & $0$ & $1$ & $2$ & $3$  & $4$  \\
\hline
\multirow{3}{*}{Subwords} & th & w-uh-d & ah-g-ow & ih-t-ae-l-y-ah-n & ih-m-ow-sh-ah-n-ah-l \\
& r  & l-ay-k & r-ay-t-ih-ng & w-eh-n-eh-v-er & ah-m-eh-r-ih-k-ah-n-z \\
& k-r & ih-n & n-eh-v-er & p-r-ay-m-eh-r-iy &  ih-f-eh-k-t-ih-v-l-iy \\
\bottomrule
\end{tabular}
 \caption{Subword units learnt by BPE in English pronunciation space.}
   \label{tab:english-bpe}
 \end{center}
\end{table*}

\begin{algorithm}
\caption{A Modified BPE Algorithm to Learn Syllables}
\label{ag1}
\begin{algorithmic}
\STATE Inputs: pronunciation sentences,  and the number of merge, $m$
\STATE $i = 0$
\WHILE{$i<m$}
\STATE $1)$ Count the frequencies of pair of symbols
\STATE  $2)$ Select the most frequent pair of symbols, with one symbol contains one vowel, and the other symbol contains no vowel
\STATE $3)$ Scan pronunciation sentences and replace all occurrences of the selected pair by a new symbol
\STATE $4)$ $i = i + 1$
\ENDWHILE
\end{algorithmic}
\end{algorithm}

BPE algorithm is a data-driven approach and the core idea of BPE is to iteratively replace the most common pair of items by a single item. To get syllables, we add a constraint on the selected pair: one of it must contain a vowel and the other one contains no vowel. Due to this constraint, the new added item is guaranteed to have one and only one vowel. Algorithm \ref{ag1} illustrates the whole algorithm. The subwords learnt in this way is not traditional syllables, but very similar. Table \ref{tab:syllables} demonstrates some syllables learnt by this algorithm. Clearly, they are very meaningful in pronunciation. How many syllable learnt in this way are well-known syllables? There is a website \footnote{\url{http://web.archive.org/web/20160822211027/http://semarch.linguistics.fas.nyu.edu/barker/Syllables/index.txt}} which lists $15831$ syllable candidates and it claims that ``there is a good chance that most, though probably not quite all legitimate syllables are represented''. Even the set of phonemes we use is a little different from the set of phonemes in these candidates, among $10004$ syllables learnt by our algorithm, $6761$ are in this list.  

\begin{table*}[h]
\begin{center}
\begin{tabular}{l|ccccc} \toprule
\#phonemes & $1$ & $2$ & $3$& $4$ & $5$ \\
\hline
\multirow{3}{*}{Syllables} &
er & t-uw & ah-n-d & m-ah-n-t & s-t-ey-t-s \\
& ao & ow-n & k-aa-n & ih-n-s-t &  t-r-ae-n-s \\
& iy & l-iy & m-ah-n & s-ih-k-s & t-w-eh-l-v \\
\bottomrule
\end{tabular}
 \caption{Syllables learnt by the modified BPE algorithm in English pronunciation space.}
   \label{tab:syllables}
 \end{center}
\end{table*}

\section{Model}
Since the state-of-the-art model for text translation is transformer \cite{vaswani2017attention}, we use exactly the same transformer networks for all our experiments. The network has $6$ transformer layers. In each layer, there are $2048$ neurons in the feed-forward part and $512$ neurons in other parts, and the number of heads for multi-head attention is $8$. For all dropout layers, the dropout rate is fixed at $0.1$. Label smoothing is used and the smoothing rate is $0.1$. Adam optimizer \cite{kingma2014adam} is used with NOAM decay and the initial learning rate is $2$. All models are trained on a computer with $8$ GPUs.

\section{Experiments}
\subsection{Basic Units}
Since subword unit is proven to be very effective for text translation. For the text sentences, we always use subwords learnt by \emph{sentencepiece} as basic units. For Chinese texts, we use about $16000$ subwords, and for English texts, we use about $10000$ subwords.

For pronunciation sentences, we experiment different basic units. For Chinese, since Pinyins can be considered to be syllables \cite{du2017pinyin}, we use both Pinyins and subwords learnt by \emph{sentencepiece} (many of them contain more than 1 Pinyins) as basic units. For English, the smallest basic units are phonemes, we also use subwords learnt by \emph{sentencepiece}, and the syllables learnt by Algorithm \ref{ag1} as basic units.  To make the comparison fair, we set the number of subwords for Chinese to be $16000$, and the number of subwords and syllables for English to be $10000$. As for Pinyins and phonemes, there are $1485$ Pinyins and $39$ phonemes.

We first compare different basic units in pronunciation space, on the task of P2P-Tran. Specially, we compare $3$ types of P2P-Tran, namely, $1)$ Pinyins for Chinese and phonemes for English, $2)$ Pinyins for Chinese and syllables for English, and $3)$ subwords for Chinese and subwords for English. We trained the models for both Zh-En and En-Zh, and for fair comparison, the translations are first transformed into sequences of Pinyins for Chinese and sequences of phonemes for English, then the $4$-gram BLEU score \cite{papineni2002bleu} is calculated by \emph{multi-bleu.perl}. Note that in this paper, all BLEU scores in the pronunciation space are calculated in this way, and all BLEU scores are single-reference scores. Table \ref{tab:basic_units} shows the results on both \emph{dev} and \emph{test} datasets. Consistent with the text translation, the subword approach (type $3$ of P2P-Tran) achieves the best performance in both directions. The syllable approach (type $2$ of P2P-Tran) is about $1$ BLEU point worse. However, when the basic unit is switched to phonemes in English side, the performance is much worse, especially in the direction of Zh-En. 

Since the performance of using phonemes as basic units in English is pretty bad, in the following experiments, we only use syllables and subwords as basic units.

\begin{table}[h]
\begin{center}
\scalebox{0.9}{
\begin{tabular}{l|c|cc} \toprule
Basic Units & Direction & Dev& Test \\
\hline
\multirow{2}{*}{\shortstack[l]{Zh: Pinyin ($1485$)\\En: phoneme ($39$)}} & Zh-En & $51.17$ & $51.75$ \\
& En-Zh & $46.39$ & $46.75$ \\
\hline
\multirow{2}{*}{\shortstack[l]{Zh: Pinyin ($1485$) \\En: syllable($10004$)}} & Zh-En & $70.15$ & $70.34$ \\
& En-Zh & $49.41$ & $49.98$ \\
\hline
\multirow{2}{*}{\shortstack[l]{Zh: subword ($16000$)\\En: subword ($10000$)}} & Zh-En & $71.06$ & $\mathbf{71.23}$\\
& En-Zh & $50.57$ & $\mathbf{50.90}$ \\
\bottomrule
\end{tabular}
}
 \caption{P2P-Tran with different basic units. The numbers in bracket in first column are the number of basic units. }
   \label{tab:basic_units}
 \end{center}
\end{table}

\subsection{T2P-Conv and P2T-Conv}
To compare two translation systems whose outputs are in different spaces, we need to convert a text sentence into a pronunciation sentence (T2P-Conv), or covert a pronunciation sentence into a text sentence (P2T-Conv). Since a pronunciation sentence usually contains less information than its textual counterpart due to the lack of punctuations, P2T-Conv is much harder than T2P-Conv.

We train models to do conversions in both directions, that is, P2T-Conv and T2P-Conv, for both Chinese and English, and the results are shown in Table \ref{tab:conversion}. For T2P-Conv, the results are almost perfect (BLEU scores are close to $100$) since this task is almost a lexicon lookup problem \footnote{The output may contains some unknown words since we use subwords as basic units for text, and the concatenation of some subwords may create unknown words.}. For P2T-Conv, the results are also pretty good,  considering that models need to predict punctuations, whose information does not directly exist in the pronunciation sentences. From this table, we can see that subwords seems to be a little better than syllables in this task.

\begin{table*}[h]
\begin{center}
\begin{tabular}{l|cc|cc} \toprule
Direction & Language & Basic Unit & Dev& Test \\
\hline
\multirow{4}{*}{T2P-Conv} & Zh & Pinyin & $99.90$ & $99.93$ \\
& Zh & subword & $99.94$ & $99.96$ \\
& En & syllable & $99.53$ & $99.65$ \\
& En & subword & $99.99$ & $99.99$ \\
\hline
\multirow{4}{*}{P2T-Conv} & Zh & Pinyin & $90.01$ & $89.83$\\
& Zh & subword & $92.10$ & $91.92$ \\
& En & syllable & $93.62$ & $93.60$ \\
& En & subword & $94.11$ & $94.21$ \\
\bottomrule
\end{tabular}
 \caption{Performance of conversion between text sentences and pronunciation sentences. Here `basic unit' refers to basic units in the pronunciation space. For text sentences, we alway use subwords as basic units.}
   \label{tab:conversion}
 \end{center}
\end{table*}

With models of both T2P-Conv and P2T-Conv, we can do conversion between text sentences and pronunciation sentences, which makes the comparison in the same space possible. However, we need to remember that the errors in T2P-Conv and P2T-Conv are different: there is only a very small amount of error in T2P-Conv, but the errors in P2T-Conv is much larger.

\subsection{P2P-Tran}
In this subsection, we report results of P2P-Tran, and compare it against T2T-Tran. Since the outputs of P2P-Tran and T2T-Tran are in different spaces, for the purpose of comparison, we report performances of all models in both spaces. To calculate BLEU scores for outputs of P2P-Tran in text space, we convert the outputs of P2P-Tran into text sentences by P2T-Conv; in the same way, to calculate BLEU scores in pronunciation space, we also convert the outputs of T2T-Tran into pronunciation sentences by T2P-Conv.

For P2P-Tran, we experiments two variants, namely, P2P-Tran-Syllable and P2P-Tran-Subword. In P2P-Tran-Syllable,  the basic unit is Pinyin for Chinese and syllable for English. In P2P-Tran-Subword, 
the basic unit is subword for both Chinese and English. Table \ref{tab:pron2pron} demonstrates performances of all models.  There are some important observations. First, the performances of all models are pretty good. This is probably because our dataset is large and the distributions of \emph{dev} and \emph{test} sets are similar to the distribution of \emph{train} set. Second, the performance of P2P-Tran-Subword is consistently a little better than P2P-Tran-Syllable, which suggests that subwords are also very good basic units for translation in pronunciation space. Second, the performance of P2P-Tran-Subword is in general comparable to the performance of T2T-Tran. In Pronunciation space, P2P-Tran-Subword is slightly better than T2T-Tran ($71.23$ vs $71.06$) in the direction of Zh-En, and T2T-Tran is slightly better than P2P-Tran-Subword ($51.02$ vs $50.90$) in the direction of En-Zh. In Text space, in both directions, T2T-Tran seems to be a little better than P2P-Tran-Subword ($43.46$ vs $42.86$ and $37.13$ vs $35.64$). However, there are errors in P2T-Conv, and when taking this into account, the differences will be subtle. Another important factor which also affects the performance of P2P-Tran is that a pronunciation sentence usually contains less information than its textual counterpart.

\begin{table*}[ht]
\begin{center}
\begin{tabular}{c|c|c|cc|cc} \toprule
Direction & Category & Basic Units & \multicolumn{2}{c}{Text Space} & \multicolumn{2}{c}{Pronunciation Space} \\
& & & Dev & Test & Dev & Test \\
\hline
\multirow{3}{*}{Zh-En} & P2P-Tran & Zh: Pinyin, En: syllable & $40.24$ & $40.87$ & $70.15$ & $70.34$ \\
& P2P-Tran & Zh: subword, En: subword & $42.36$ & $42.86$ & $71.06$ & $\mathbf{71.23}$ \\
& T2T-Tran & Zh: subword, En: subword & $43.16$ & $\mathbf{43.46}$ & $70.91$ & $71.06$ \\
\hline
\multirow{3}{*}{En-Zh} & P2P-Tran & Zh: Pinyin, En: syllable & $33.19$ & $34.21$ & $49.41$ & $49.98$ \\
& P2P-Tran & Zh: subword, En: subword & $34.79$ & $35.64$ & $50.57$ & $50.90$ \\
& T2T-Tran & Zh: subword, En: subword & $36.26$ & $\mathbf{37.13}$ & $50.56$ & $\mathbf{51.02}$ \\
\bottomrule
\end{tabular}
 \caption{Performances of P2P-Tran, compared with T2T-Tran}
   \label{tab:pron2pron}
 \end{center}
\end{table*}

\subsection{P2T-Tran}
In this subsection, we report results on P2T-Tran, and also compare it against T2T-Tran. Since the outputs of both P2T-Tran and T2T-Tran are in text space, we only report their performances in text space. 
For P2T-Tran,  we also experiment two variants, namely, P2T-Tran-Syllable and P2T-Tran-Subword. The results are illustrated in Table \ref{tab:pron2text}. On the test set, the performance difference between
P2T-Tran-Syllable and P2T-Tran-Subword is very small and almost neglect-able, which is different from the results on P2P-Tran, where differences between P2P-Tran-Syllable and P2P-Tran-Subword is small but noticeable. 
From Table \ref{tab:pron2text}, we also see that the performance of P2T-Tran-Subword is about $1$ BLEU worse than the performance of T2T-Tran, probably because the fact that a pronunciation sentence contains less information than their corresponding text sentence.

\begin{table*}[ht]
\begin{center}
\begin{tabular}{c|c|c|cc} \toprule
Direction & Category & Basic Units & \multicolumn{2}{c}{Text Space}  \\
& & & Dev & Test  \\
\hline
\multirow{3}{*}{Zh-En} & P2T-Tran & Zh: Pinyin, En: subword & $41.93$ & $42.51$ \\
& P2T-Tran & Zh: subword, En: subword & $42.59$ & $42.75$  \\
& T2T-Tran & Zh: subword, En: subword & $43.16$ & $\mathbf{43.46}$  \\
\hline
\multirow{3}{*}{En-Zh} & P2T-Tran & En: syllable, Zh: subword & $35.12$ & $36.12$  \\
& P2T-Tran & En: subword, Zh: subword & $35.25$ & $36.04$  \\
& T2T-Tran & En: subword, Zh: subword & $36.26$ & $\mathbf{37.13}$  \\
\bottomrule
\end{tabular}
 \caption{Performance of P2T-Tran}
   \label{tab:pron2text}
 \end{center}
\end{table*}

\subsection{T2P-Tran}
In this subsection, we report results on T2P-Tran. Since the outputs are all in pronunciation space, we only report BLEU scores in pronunciation space. We also experiment two variants of T2P-Tran namely, T2P-Tran-Syllable and T2P-Tran-Subword. The results are demonstrated in Table \ref{tab:text2pron}. Since the inputs of both T2P-Tran and P2P-Tran are text sentences, there is no difference in information in source side. Although we need to convert the output of T2T-Tran into pronunciation sentences by T2P-Conv, the loss in this process is tiny according to Table \ref{tab:conversion}. From Table \ref{tab:text2pron}, we can see that the performances of these 3 translation systems are almost the same, with very small differences in their BLEU scores.

\begin{table*}[ht]
\begin{center}
\begin{tabular}{c|c|c|cc} \toprule
Direction & Category & Basic Units & \multicolumn{2}{c}{Pronunciation Space}  \\
& & & Dev & Test  \\
\hline
\multirow{3}{*}{Zh-En} & T2P-Tran & Zh: subword, En: syllable & $71.48$ & $71.43$ \\
& T2P-Tran & Zh: subword, En: subword & $71.65$ & $\mathbf{71.75}$  \\
& T2T-Tran & Zh: subword, En: subword & $70.91$ & $71.06$  \\
\hline
\multirow{3}{*}{En-Zh} & T2P-Tran & En: subword, Zh: Pinyin & $50.35$ & $50.74$  \\
& T2P-Tran & En: subword, Zh: subword & $50.73$ & $\mathbf{51.09}$  \\
& T2T-Tran & En: subword, Zh: subword & $50.56$ & $51.02$  \\
\bottomrule
\end{tabular}
 \caption{Performance of T2P-Tran}
   \label{tab:text2pron}
 \end{center}
\end{table*}

\section{Related Work}
Our work is closely related to text translation, and there are tons of research literatures on text translation for various languages \cite{brown1990statistical, koehn2009statistical, bahdanau2014neural, luong2015effective, wu2016google}. As verified by our experiments, many methods suitable for text translation is also good for translation in pronunciation space, such as transformer network \cite{vaswani2017attention}, the subword approach \cite{sennrich2016neural}. From the perspective of deep learning, both of them are in essence sequence-to-sequence problems \cite{sutskever2014sequence}. However, there is also a lot of differences. For example, the pronunciation units, such as Pinyins and phonemes, are usually more limited and constrained than words in many languages. Many activities, such as speech recognition \cite{povey2011kaldi} and Text-to-Speech \cite{arik2017deep}, inherently occur in pronunciation space.

There is also some previous work on Grapheme-to-Phoneme  \cite{taylor2005hidden, bisani2008joint}. Before the era of neural network, the most important approach on G2P is joint-sequence models \cite{bisani2008joint}. Recently, there are a lot of methods based on neural networks, especially the seq2seq models \cite{rao2015grapheme, deri2016grapheme}. Many of these approaches are based on a single word, that is, producing the pronunciation of this word without taking the context into consideration. Our T2P-Conv is based on the whole sentences and thus utilizes the context.

\section{Conclusion}
In this paper, we conducted large scale experiments on three new categories of translations, namely, P2P-Tran, P2T-Tran and T2P-Tran, and found that their performances are on par with the performance of T2T-Tran, the traditional machine translation.
Since no existing datasets, we build a large dataset based on WMT18 Zh-En dataset. We also proposed a modified BPE algorithm to learn syllables, and experimented different basic units, such as phonemes, Pinyins, syllables and subwords, for pronunciation sentences. We found that similar to text sentences, subword is in general best choice for pronunciation sentences. However, syllable is good meaningful alternative. Due to similar performance of P2P-Tran, P2T-Tran, T2P-Tran and T2T-Tran, we advocate that a systems should choose the one, among these four categories of translations, which can simplify the whole system, according to its upstream and downstream components. 

\clearpage

\bibliography{pronMT}
\bibliographystyle{acl_natbib}

\end{CJK}

\end{document}